\documentclass[11pt]{article}
\usepackage{graphicx}
\usepackage[dvipsnames]{xcolor}
\usepackage{geometry}
\usepackage{hyperref}
\hypersetup{
  colorlinks   = true,
  urlcolor     = blue,
  linkcolor    = blue,
  citecolor    = black
}
\usepackage[utf8]{inputenc}
\usepackage{natbib} 
\usepackage{booktabs} 
\usepackage{siunitx}
\usepackage{gensymb}
\usepackage{setspace}
\usepackage{lineno}
\usepackage{wrapfig}
\usepackage{authblk}
\usepackage{cleveref}

\title{From images to properties: a NeRF-driven framework for granular material parameter inversion}

\author[1]{Cheng-Hsi Hsiao}
\author[1]{Krishna Kumar}

\affil[1]{Maseeh Department of Civil, Architectural and Environmental Engineering,\newline
The University of Texas at Austin, Austin, TX, USA}

\begin{document}

\maketitle


\begin{abstract}
We introduce a novel framework that integrates Neural Radiance Fields (NeRF) with Material Point Method (MPM) simulation to infer granular material properties from visual observations. Our approach begins by generating synthetic experimental data, simulating an plow interacting with sand. The experiment is rendered into realistic images as the photographic observations. These observations include multi-view images of the experiment’s initial state and time-sequenced images from two fixed cameras. Using NeRF, we reconstruct the 3D geometry from the initial multi-view images, leveraging its capability to synthesize novel viewpoints and capture intricate surface details. The reconstructed geometry is then used to initialize material point positions for the MPM simulation, where the friction angle remains unknown. We render images of the simulation under the same camera setup and compare them to the observed images. By employing Bayesian optimization, we minimize the image loss to estimate the best-fitting friction angle. Our results demonstrate that friction angle can be estimated with an error within 2\degree, highlighting the effectiveness of inverse analysis through purely visual observations. This approach offers a promising solution for characterizing granular materials in real-world scenarios where direct measurement is impractical or impossible.
\end{abstract}

\section{Introduction}
Soil properties, particularly the internal friction angle, are critical parameters in geotechnical engineering, governing the behavior of slopes, foundations, retaining structures, and excavation processes. These properties are typically measured through laboratory experiments such as direct shear or triaxial tests. However, such tests are impractical or impossible in remote environments like the Moon or Mars, where extracting soil samples and transporting specialized equipment pose significant logistical and financial challenges.

Apollo mission data suggest that the friction angle of lunar soil ranges from 30\degree to 50\degree, depending on soil density and location~\citep{mitchell_mechanical_1972}. However, most of these values were derived from limited sampling in lunar mare regions, making it difficult to generalize across diverse terrain. As space agencies prepare for more ambitious missions, there is a pressing need to develop alternative, in-situ methods to estimate soil properties using only visual sensors mounted on planetary rovers.

At the same time, advances in computer vision have introduced new tools for recovering 3D geometry from images. Neural Radiance Fields (NeRF, \citealt{mildenhall_nerf_2020}) have significantly improved novel view synthesis and 3D reconstruction from sparse multi-view imagery. This method produces photorealistic scene representations with high geometric fidelity.
This capability opens up new possibilities for material characterization in scenarios where direct measurement is infeasible. For example, in a real mission scenario, a planetary rover could perform a controlled soil interaction, such as dragging a plow across the surface. Before the interaction, the rover collects multi-view images of the undisturbed terrain. As the plow moves through the soil, it deforms the surface, and this dynamic interaction is recorded over time by the fixed cameras.

\begin{figure}[h]
\centering
\includegraphics[width=0.7\linewidth]{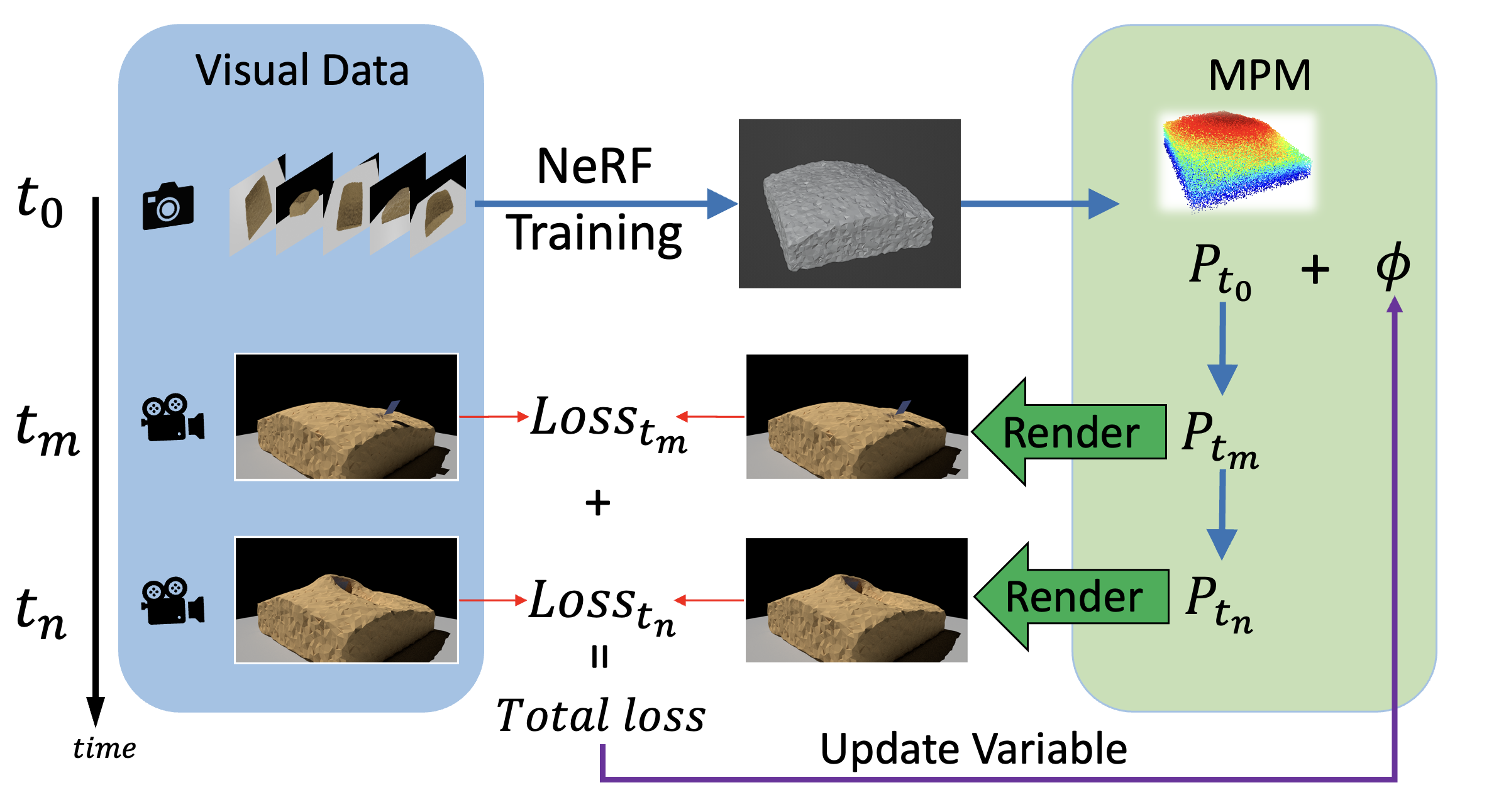}
\caption{Image to property framework. The undeformed geometry at time $t_0$ is reconstructed from multi-view images. Initial material points $p_{t_0}$ are sampled from the reconstructed geometry and assigned a candidate material property $\phi$. The simulated material state $p_{t_m}$ at time $t_m$ is rendered and compared to the observed images. Image losses across multiple time steps are summed and used to update the candidate property for the next simulation.
}
\label{framework}       
\end{figure}

Our framework uses this visual data to infer the soil's friction angle, as illustrated in~\Cref{framework}. First, the initial multi-view images are used to reconstruct the 3D geometry of the terrain using NeRF. This reconstruction provides the initial condition for a MPM simulation. We then simulate the same plow-soil interaction using various candidate friction angles, rendering the deformed material from the same camera viewpoints as the observations. By computing the image loss between the simulated and observed sequences, we apply Bayesian Optimization to identify the friction angle that best reproduces the visual evidence of the interaction

Our contributions are as follows:
\begin{itemize}
    \item We introduce a NeRF-MPM framework for granular material property estimation from images
    \item We apply Bayesian optimization to inverse the friction angle by minimizing image loss.
    \item We validate our framework using synthetic plow-sand interactions across multiple ground truth friction angles, achieving mean absolute errors between 0.64\degree{} and 1.38\degree{}, demonstrating the feasibility and robustness of vision-based soil property inference.
\end{itemize}

This method opens up a new direction for in-situ geotechnical characterization using only visual data. It has potential for applications in planetary exploration and other field scenarios where direct sampling is infeasible, and only visual data is available.

\section{State of the Art}
\label{sec:related_work}

\subsection{Novel View Synthesis and Scene Reconstruction}

NeRF~\citep{mildenhall_nerf_2020} represents scenes as continuous volumetric fields, enabling high-fidelity novel view synthesis that captures fine surface details. Subsequent advancements such as Mip-NeRF~\citep{barron_mip-nerf_2021} and Instant-NGP~\citep{muller_instant_2022} improve rendering quality and training speed by addressing aliasing and computational efficiency. More recently, Gaussian Splatting~\citep{kerbl_3d_2023} introduces a fast, non-neural alternative that represents scenes using rasterized 3D Gaussians for real-time performance.

Recent work has extended NeRF to incorporate physical modeling for simulation-aware reconstruction. PAC-NeRF~\citep{li_pac-nerf_2023} jointly estimates both geometry and material parameters by coupling NeRF with a differentiable MPM solver, using a hybrid Eulerian-Lagrangian representation to enforce physical plausibility. PIE-NeRF~\citep{feng_pie-nerf_2023} introduces a meshless hyperelastic simulation framework that operates directly on NeRF-reconstructed geometries, enabling efficient modeling of large deformations without traditional mesh discretization.

While these approaches tightly couple physical simulation and rendering through differentiable pipelines, our method separates the two: we use NeRF solely for geometry reconstruction and perform parameter inversion using Bayesian Optimization, without requiring simulator gradients. This decoupled design allows our framework to work seamlessly with existing simulation tools, including those that are non-differentiable or closed-source.

\section{Methodology}
\label{sec-method}
\subsection{Neural Radiance Fields (NeRF)}
\label{sec-method-nerf}
NeRF is a neural network-based technique for view synthesis and 3D scene reconstruction from 2D images. It learns an implicit representation of the underlying 3D structure.

\begin{figure}[h]
\centering
\includegraphics[width=0.7\linewidth]{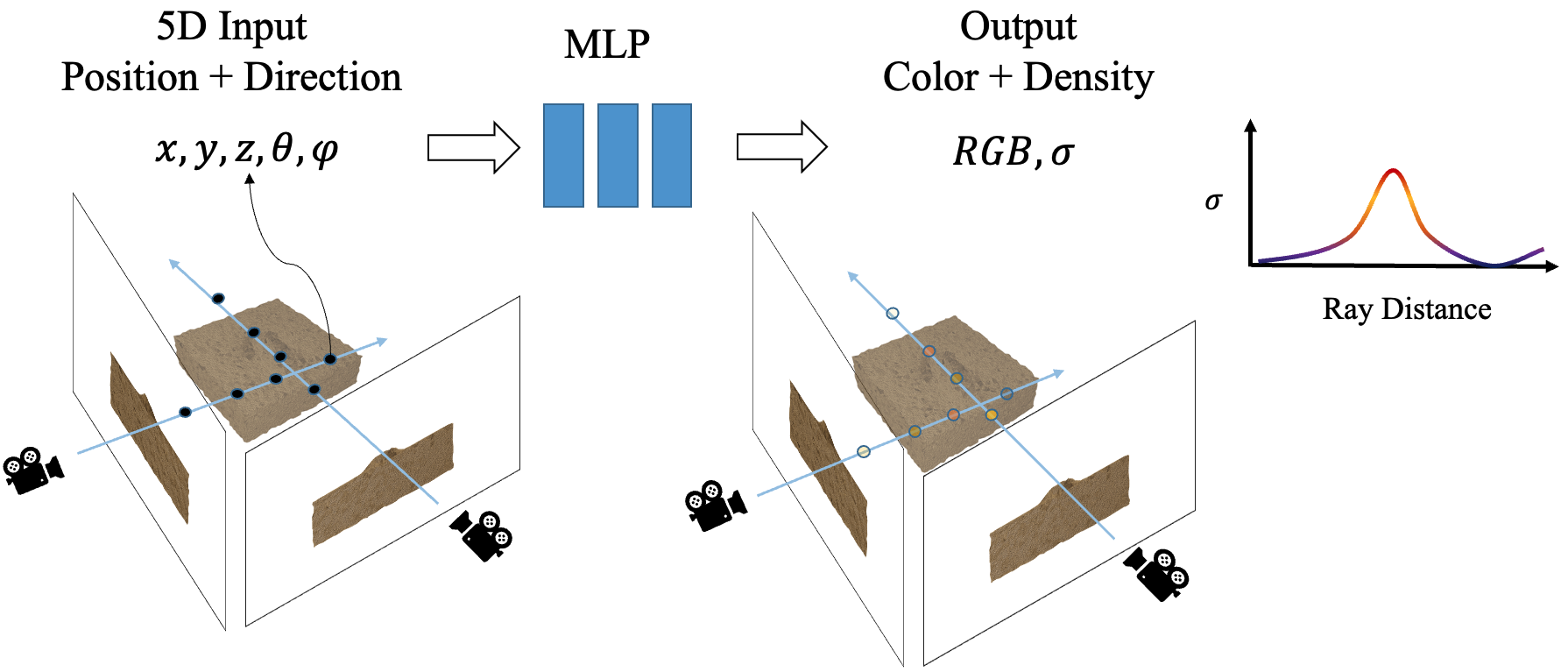}
\caption{Overview of Neural Radiance Field}
\label{nerf}       
\end{figure}

NeRF represents a scene as a continuous volumetric field that maps 3D spatial coordinates ($xyz$) and viewing directions ($\theta$ and $\varphi$) to color (RGB) and volume density $\sigma$ (see in~\Cref{nerf}). This is achieved using a multi-layer perceptron (MLP) that learns to approximate the radiance field of the scene. Given a set of input images with known camera poses, NeRF is trained to optimize a rendering loss, ensuring that synthesized images from novel viewpoints match the observed images. The model effectively captures fine-grained geometric and appearance details, making it well-suited for photo-realistic 3D reconstruction.

\subsection{Bayesian Optimization}
\label{sec-method-bo}
Bayesian Optimization (BO) is used to estimate the friction angle by minimizing the image loss between simulated and observed scenes. Since each simulation is computationally expensive, BO builds a surrogate model to guide the search efficiently.

The process begins by evaluating one or a few randomly selected friction angles and recording their image losses. A Gaussian Process (GP) is then fit to this data, modeling both the predicted loss and the uncertainty across the search space. To select the next angle to try, we use the Expected Improvement (EI) acquisition function, which favors candidates that are likely to improve over the current best result. EI balances exploring uncertain regions and refining promising areas, making it suitable for low-budget optimization.

After each new simulation, the GP model is updated with the result, and the process repeats until the evaluation budget is reached or the loss converges.

\section{Results}
\label{result}

In this study, the observation data is generated from MPM simulations with sand friction angles of 25\degree, 30\degree, 35\degree, and 45\degree. Each simulation runs for 200 frames with a time step of 0.02 seconds per frame. The simulated plow interaction consists of four stages: (1) the plow is lowered into the sand over the first 25 frames; (2) it pushes forward for the next 75 frames; (3) the system pauses for 25 frames; and (4) the plow is pulled back during the final 75 frames. All simulations are executed on an NVIDIA A100 GPU and complete in approximately 5 minutes.

At the initial frame, we render 40 multi-view images from different camera positions to serve as input for NeRF training. Additionally, two fixed cameras are used to capture images at each time step throughout the simulation. These multi-view and time-sequenced images serve as the observational dataset used to identify the friction angle.

We train the NeRF model using the 40 multi-view images to reconstruct the 3D geometry of the sand's initial state. The training process takes approximately one hour on an A100 GPU. Material points for subsequent simulations are sampled from the reconstructed NeRF geometry. We then rerun the MPM simulation using various candidate friction angles, applying the same plow trajectory as in the observation case. The simulations are recorded using the same fixed camera settings, and we compare the resulting images to the ground truth at frames 50, 100, 150, and 199. The comparison is quantified by computing the image loss, which is the mean square error of the RGB channels.

\begin{figure*}[h]
\centering
\vspace*{1cm}       
\includegraphics[width=0.9\textwidth]{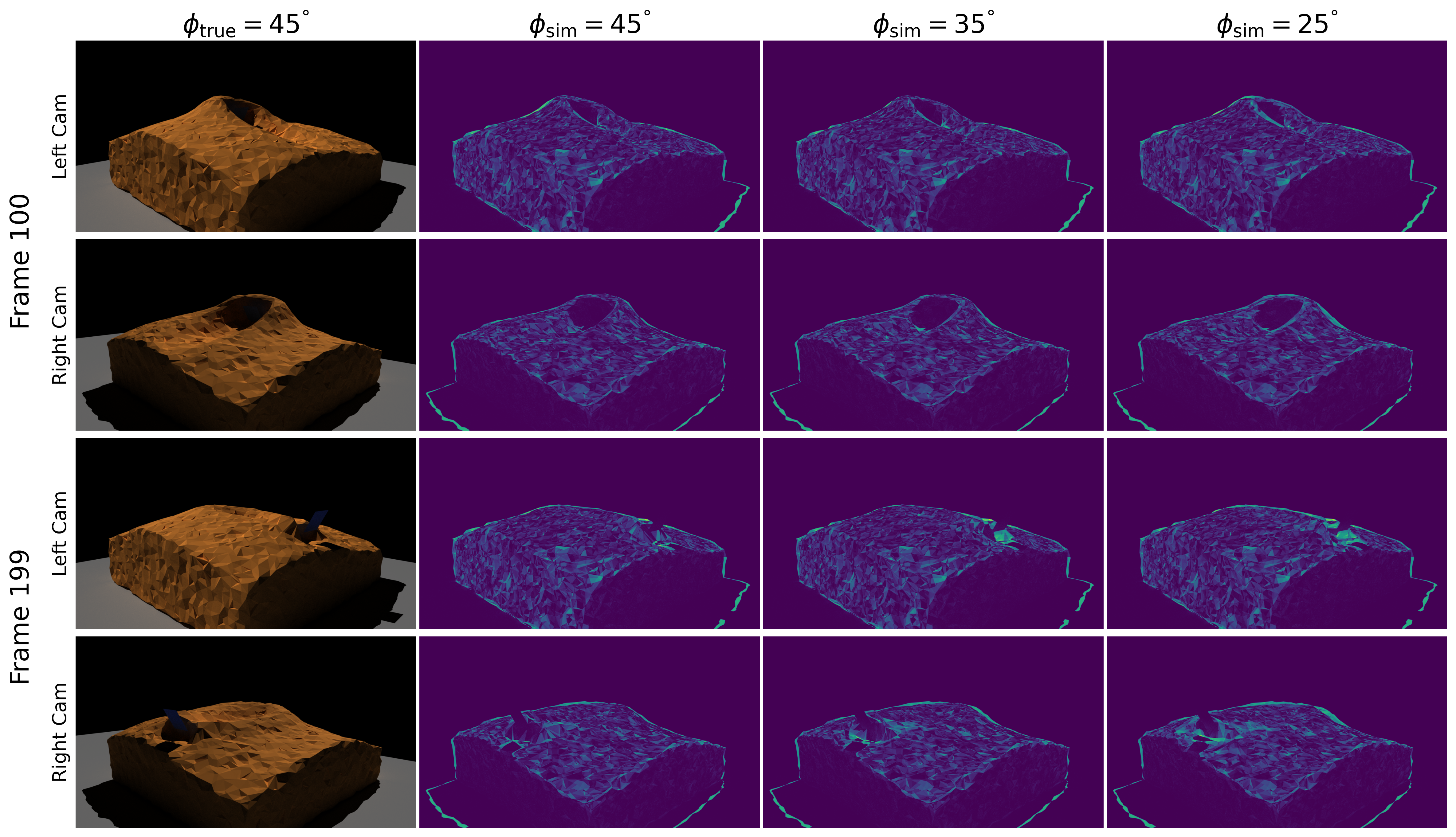}
\caption{Image comparison between observed and simulated scenes at frames 100 and 199 from both left and right camera views. The first column shows the observed images from the ground truth simulation with $\phi_{true} = 45\degree$. The second, third, and fourth columns display the pixel-wise image differences between the observation and simulations with friction angles $\phi_{sim} = 45\degree$, $35\degree$, and $25\degree$, respectively. Differences are most pronounced near the crown and behind the plow, especially as $\phi_{sim}$ deviates further from $\phi_{true}$.}
\label{fig:comparison-45}       
\end{figure*}

\Cref{fig:comparison-45} presents a visual comparison between the observed and simulated images for the case where the ground truth friction angle is 45\degree. The first column shows the reference images from the original simulation (experiment). The top two rows correspond to the state after the plow finishes pushing forward. We observe that the sand has deformed, forming a trench in the plow’s path and a pile around the plow due to material accumulation. The bottom two rows show the state after the plow is pulled back. At this stage, the unsupported pile collapses, and additional material accumulates behind the plow as it moves backward.

The second to fourth columns of~\Cref{fig:comparison-45} show the difference heat map for simulations with friction angles $\phi_{sim} = 45\degree$, $35\degree$, and $25\degree$. Even when the simulated friction angle matches the ground truth ($\phi_{sim} = \phi_{true} = 45\degree$), image differences are still present. This is due to differences in initial geometry: the material points for simulation are sampled from the NeRF-reconstructed geometry, which is not identical to the original simulation. This discrepancy introduces background noise in all comparisons.

Despite this, we observe that the image differences increase as the simulated friction angle deviates further from the ground truth. This effect is especially visible around the plow. For example, in frame 100, larger discrepancies appear at the top of the crown and along the trailing edge of the pile. In frame 199, greater image differences are observed behind the plow, where sand accumulates during the backward motion. These results confirm that image loss is sensitive to changes in the friction angle and is therefore a viable objective for optimization.

\begin{figure}[h]
\centering
\includegraphics[width=0.5\linewidth]{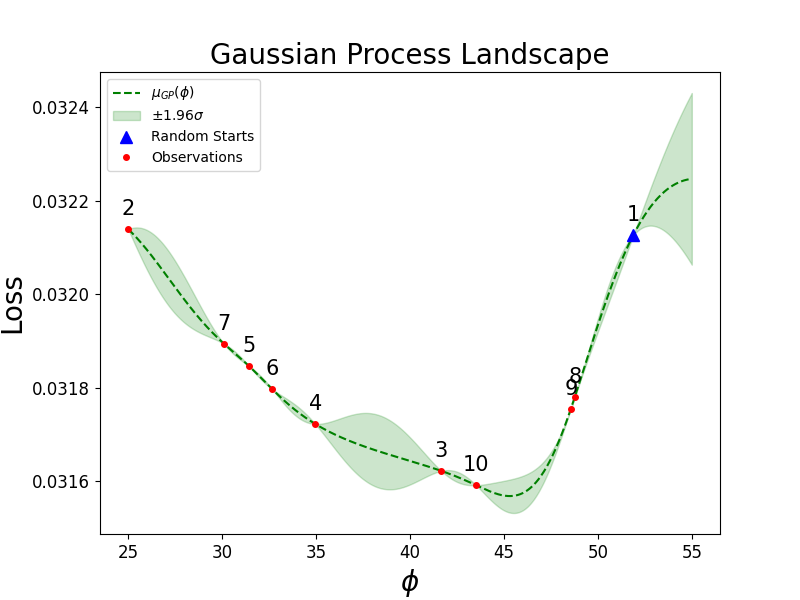}
\caption{Bayesian optimization result for $\phi_{true}=45\degree$.}
\label{fig:bo_result}       
\end{figure}

For Bayesian Optimization, we limit the total number of evaluations to 10 due to the computational cost of the simulations.~\Cref{fig:bo_result} illustrates the Bayesian Optimization process for $\phi_{true}=45\degree$. The triangle marker represents the initial observation, which is randomly sampled from a range of 25\degree to 55\degree. The red dots denote subsequent evaluations which are selected based on the updated GP model. The dashed line represents the mean prediction of the GP model, while the green region indicates the uncertainty interval. After 10 evaluations (including initial observations), the optimized friction angle is 43.58\degree.

We investigate the stability of Bayesian Optimization by performing 10 independent trials for each ground truth friction angle. Each trial is allowed a maximum of 10 evaluations.~\Cref{tab:bo_result} reports the mean absolute error (MAE) and standard deviation for each case. The MAE ranges from 0.64\degree to 1.38\degree, demonstrating that the optimization is generally accurate and robust. However, we observe that both the error and variance tend to increase with higher ground truth friction angles. This suggests that the image loss landscape becomes more challenging to optimize as the friction angle increases, potentially due to more subtle deformation patterns that are harder to distinguish through visual comparison.

\begin{table}[h]
\centering
\caption{Mean absolute error and its standard deviation across 10 Bayesian Optimization trials.}
\label{tab:bo_result}
\begin{tabular}{cc}
\toprule
\textbf{Ground Truth (°)} & \textbf{Error (MAE ± Std) (°)} \\
\midrule
25 & 0.64 ± 1.01 \\
30 & 1.07 ± 0.71 \\
35 & 1.24 ± 0.96 \\
45 & 1.38 ± 1.36 \\
\bottomrule
\end{tabular}
\end{table}

\section{Conclusion}
\label{conclusion}
We presented a novel framework that combines NeRF reconstruction with MPM simulations to infer granular material properties from visual data. Using Bayesian optimization to minimize image loss between observed and simulated frames, we demonstrated the potential for accurate friction angle estimation with purely visual input. In our synthetic experiments, estimated angles remained within ~2° of ground truth across multiple test cases.

This method has strong implications for field applications where direct soil testing is infeasible, such as lunar exploration. Future work will focus on testing real-world video, extending to other soil types, and improving NeRF reconstruction accuracy to further reduce error in inverse predictions.

\bibliographystyle{apalike}  
\bibliography{main}
\end{document}